\documentclass[letterpaper, 10 pt, journal, twoside]{format/IEEEtran}

\IEEEoverridecommandlockouts    
\usepackage{glossaries}
\newacronym{mav}{MAV}{Micro Aerial Vehicles}
\newacronym{uav}{UAV}{Unmanned Aerial Vehicle}
\newacronym{ovc}{OVC}{Open Vision Computer}
\newacronym{lidar}{LiDAR}{Light Detection and Ranging}
\newacronym{vio}{VIO}{visual-inertial odometry}
\newacronym{gpgpu}{GPGPU}{General-Purpose Graphics Processing Unit}
\newacronym{ugv}{UGV}{Unmanned Ground Vehicle}
\newacronym{uwb}{UWB}{Ultra Wideband}
\newacronym{svm}{SVM}{Support Vector Machine}
\newacronym{fcn}{FCN}{Fully Convolutional Network}
\newacronym{cnn}{CNN}{Convolutional Neural Network}
\newacronym{loam}{LOAM}{LiDAR Odometry and Mapping}
\newacronym{sloam}{SLOAM}{Semantic LiDAR Odometry and Mapping}
\newacronym{slam}{SLAM}{Simultaneous Localization and Mapping}
\newacronym{iot4ag}{IoT4Ag}{NSF Engineering Research Center for the Internet of Things for Precision Agriculture}
\newacronym{grasp-lab}{GRASP Lab}{the General Robotics, Automation, Sensing and Perception Laboratory}
\newacronym{jps}{JPS}{Jump Point Search}
\newacronym{ukf}{UKF}{Unscented Kalman Filter}
\newacronym{sam}{SAM}{Smoothing and Mapping}
\newacronym{icp}{ICP}{Iterative Closest Point}
\newacronym{imu}{IMU}{Inertial Measurement Unit}
\newacronym{tsdf}{TSDF}{Truncated Signed Distance Field}
\newacronym{esdf}{ESDF}{Euclidean Signed Distance Field}
\newacronym{sdf}{SDF}{Signed Distance Field}
\newacronym{rrt}{RRT}{Rapidly Exploring Random Tree}
\newacronym{fpv}{FPV}{First-person View}
\newacronym{dnn}{DNN}{Deep Neural Network}
\newacronym{igpred}{IGPred}{Information Gain Prediction}
\newacronym{csqmi}{CSQMI}{Cauchy-Schwarz Quadratic Mutual Information}
\newacronym{nbv}{NBV}{Next Best View}
\newacronym{vae}{VAE}{Variational Autoencoder}
\newacronym{tsp}{TSP}{Traveling Salesman Problem}
\newacronym{bcsm}{BCSM}{Behavior Control State Machine}
\newacronym{pca}{PCA}{Principal Component Analysis}
\newacronym{aspp}{ASPP}{Atrous Spatial Pyramid Pooling}
\newacronym{swap}{SWaP}{Size Weight and Power}
\newacronym{soi}{SoI}{Semantic Object of Interest}
\newacronym{aoi}{AoI}{Area of Interest}
\newacronym{drl}{DRL}{Deep Reinforcement Learning}
\newacronym{dl}{DL}{Deep Learning}
\newacronym{fov}{FoV}{Field of View}
\newacronym{tops}{TOPS}{Tera Operations per Second}

\usepackage[table,usenames,dvipsnames]{xcolor}      

\usepackage{tabularx}
\usepackage{multirow, multicol}
\usepackage{array}
\usepackage{tabu}
\usepackage{arydshln}
\usepackage{makecell}

\usepackage{tikz}
\usepackage{pgfplots}
\usetikzlibrary{shapes.geometric, arrows}
\usepackage{float}

\newcolumntype{P}[1]{>{\centering\arraybackslash}p{#1}}
\newcolumntype{M}[1]{>{\centering\arraybackslash}m{#1}}
\usepackage{booktabs}
\newcolumntype{N}{>{\centering\arraybackslash}m{.5in}}
\newcolumntype{G}{>{\centering\arraybackslash}m{2in}}

\newcommand\scalemath[2]{\scalebox{#1}{\mbox{\ensuremath{\displaystyle #2}}}}

\newif\ifdraft
\drafttrue %
\ifdraft
\newcommand\todo[1]{\textcolor{VioletRed}{\textbf{TODO:} #1}}
\newcommand\canremove[1]{\textcolor{blue}{#1}}

\newcommand\review[1]{}
\newcommand\WIP[1]{}
\newcommand{\VM}[1]{\textcolor{cyan}{\textbf{VM:} #1}} 
\newcommand{\DO}[1]{\textcolor{blue}{\textbf{DO:} #1}} 
\newcommand{\PC}[1]{\textcolor{red}{\textbf{PC:} #1}} 
\newcommand{\YT}[1]{\textcolor{brown}{\textbf{YT:} #1}} 
\newcommand{\vijay}[1]{\textcolor{red}{\emph{vijay:} #1}} 
\else
\newcommand\todo[1]{}
\newcommand\canremove[1]{}

\newcommand\review[1]{}
\newcommand\WIP[1]{}
\newcommand{\VM}[1]{} 
\newcommand{\DO}[1]{} 
\newcommand{\PC}[1]{} 
\newcommand{\YT}[1]{} 
\newcommand{\vijay}[1]{} 
\fi
\let\labelindent\relax

\usepackage{extarrows}                              
\usepackage{enumitem}
\usepackage{wrapfig}

\usepackage{amsmath,amssymb,amsfonts,amsthm,dsfont, bm} 
\usepackage{bbm}
\usepackage{mathtools}
\usepackage{algorithm,algorithmicx,listings}        
\usepackage[noend]{algpseudocode}			              
\makeatletter
\def\BState{\State\hskip-\ALG@thistlm}
\makeatother
\DeclarePairedDelimiter\abs{\lvert}{\rvert}%
\DeclarePairedDelimiter\norm{\lVert}{\rVert}%
\newcommand{\E}{\mathrm{E}}
\newcommand{\Var}{\mathrm{Var}}

\makeatletter
\let\oldabs\abs
\def\abs{\@ifstar{\oldabs}{\oldabs*}}
\let\oldnorm\norm
\def\norm{\@ifstar{\oldnorm}{\oldnorm*}}
\makeatother

\usepackage{graphicx}
\usepackage{subcaption}
\usepackage{textcomp}
\usepackage[font=footnotesize]{caption}
\usepackage{subcaption}

\def\argmin{\mathop{\arg\min}\limits}	%
\def\argmax{\mathop{\arg\max}\limits}	%

\usepackage{xspace}
\usepackage{placeins}

\makeatletter
\DeclareRobustCommand\onedot{\futurelet\@let@token\@onedot}
\def\@onedot{\ifx\@let@token.\else.\null\fi\xspace}

\def\ie{\emph{i.e}\onedot}

\makeatother

\usepackage[normalem]{ulem}
\usepackage[utf8]{inputenc}
\usepackage[english]{babel}

\def\argmin{\mathop{\arg\min}\limits}	%
\def\argmax{\mathop{\arg\max}\limits}	%

\DeclareMathAlphabet\mathbfcal{OMS}{cmsy}{b}{n}

\newtheorem*{assumption*}{Assumption}

\newtheorem*{problem*}{Problem}
\newtheorem{problem}{Problem}

\usepackage{graphicx}

\usepackage{lipsum}
\makeatletter
\let\NAT@parse\undefined
\makeatother
\usepackage[space, compress, sort, noadjust]{cite}
\usepackage[breaklinks=true, colorlinks, bookmarks=true, citecolor=Black, urlcolor=Violet,linkcolor=Black]{hyperref}

\begin{document}

\title{
RT-GuIDE: Real-Time Gaussian Splatting for Information-Driven Exploration}


\author{Yuezhan Tao, Dexter Ong, Varun Murali, Igor Spasojevic, Pratik Chaudhari and Vijay Kumar %
\thanks{
This work was supported by TILOS under NSF Grant CCR-2112665, IoT4Ag ERC under NSF Grant EEC-1941529, NSF grant CMMI-2415249, NSF NRI/USDA award 2022-67021-36856, the ARL DCIST CRA W911NF-17-2-0181, DSO National Laboratories and NVIDIA.}
\thanks{All authors are with GRASP Laboratory, University of Pennsylvania {\tt\footnotesize\{yztao, odexter, mvarun, igorspas, pratikac, kumar\}@seas.upenn.edu}.} %
}

\maketitle

\begin{abstract}
We propose a framework for active mapping and exploration that leverages Gaussian splatting for constructing dense maps.
Further, we develop a GPU-accelerated motion planning algorithm that can exploit the Gaussian map for real-time navigation. 
The Gaussian map constructed onboard the robot is optimized for both photometric and geometric quality while enabling real-time situational awareness for autonomy.
We show through viewpoint selection experiments that our method yields comparable Peak Signal-to-Noise Ratio (PSNR) and similar reconstruction error to state-of-the-art approaches, while being orders of magnitude faster to compute. In closed-loop physics-based simulation and real-world experiments, our algorithm achieves better map quality (at least 0.8dB higher PSNR and more than 16\% higher geometric reconstruction accuracy) than maps constructed by a state-of-the-art method, enabling semantic segmentation using off-the-shelf open-set models. Experiment videos and more details can be found on our project page:
\url{https://tyuezhan.github.io/RT_GuIDE/} 


\end{abstract}

\section{Introduction}
\label{sec:intro}
\IEEEPARstart{A}{ctive}
mapping is a problem of optimizing the trajectory of an autonomous robot in an unknown environment to construct an informative map in real-time.
It is a critical component of numerous real-world applications such as precision agriculture, infrastructure inspection, and search and rescue missions. 
While nearly all tasks rely on recovering accurate metric information to enable path planning, many also require more fine-grained information.
Recent advances in learned map representations from the computer vision and graphics communities~\cite{mildenhall2021nerf, kerbl20233dgs} have opened up new possibilities for active mapping and exploration while maintaining both geometrically and visually accurate digital twins of the environment.

While prior work effectively solves the problem of \emph{information-driven exploration}~\cite{LukasIG, alexis2020MP} or \emph{frontier-based exploration}~\cite{yamauchi1997frontier, 10143748}, in this work we consider the additional problem of generating radiance fields while also performing autonomous navigation. 
Prior work has also proposed information metrics using novel learned scene representations that are capable of high quality visual reconstruction but are incapable of running in real-time onboard a robot~\cite{he2023active}.
To enable efficient mapping and planning in these novel representations, we consider approximation techniques for computing the information gain.
Further, we consider the problem of generating high-quality maps that are capable of novel-view synthesis that can be used for downstream applications.

\begin{figure} [!t]
\centering
\includegraphics[width=\linewidth, trim={1.7cm 7.0cm 3cm 6cm},clip]{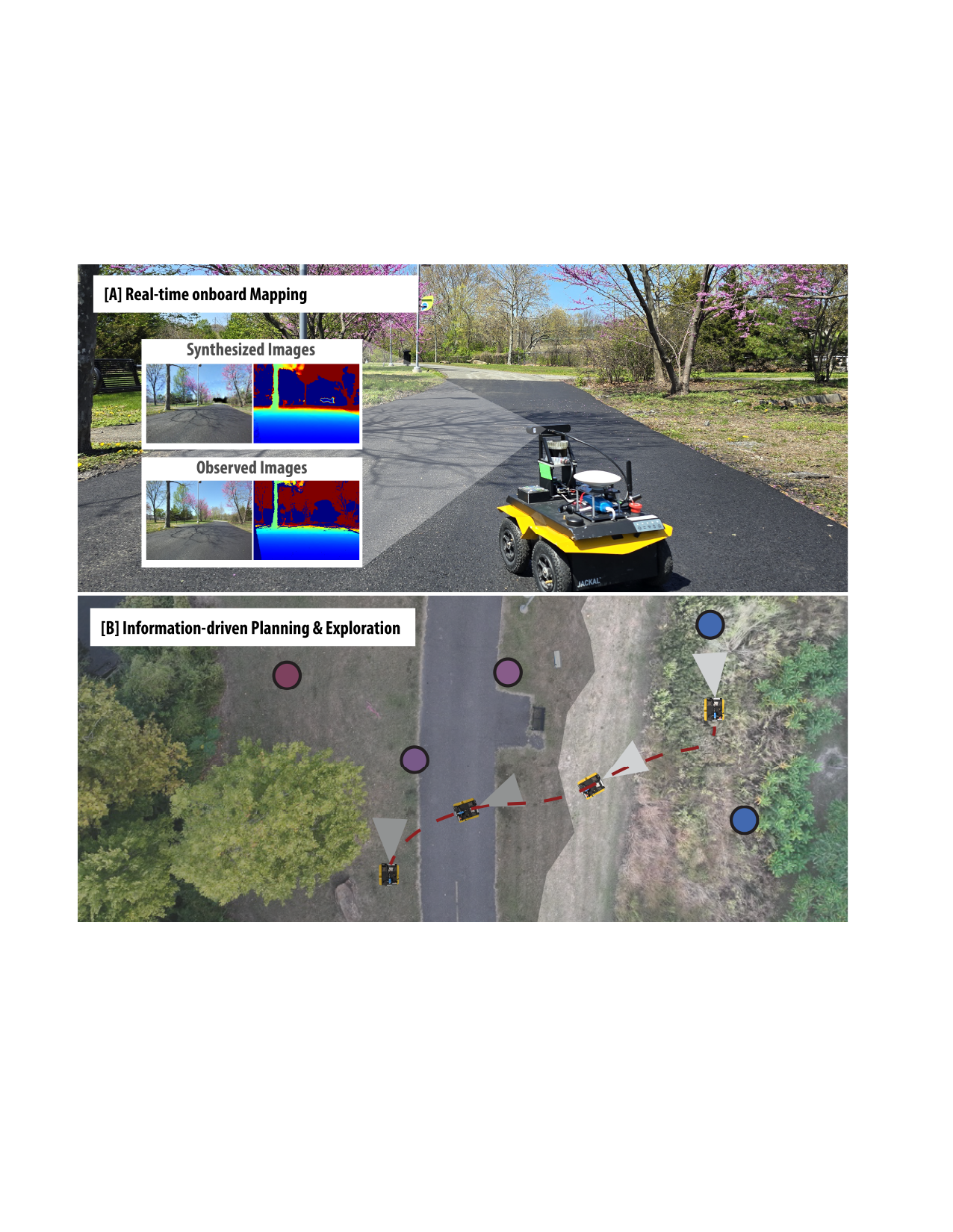}
\vspace{-0.4cm}
\caption{Key elements of our proposed approach. 
[A] Robot building a Gaussian map onboard in real-time and using it to avoid obstacles in the environment. 
Synthesized color and depth images from the Gaussian map are presented next to the corresponding observations from the RGBD sensor. 
[B] Robot navigating to unobserved areas (right) with high information gain while maximizing information along the trajectory. 
Map regions colored light to dark in increasing estimated information gain. }
\label{fig:proposed-method}
\vspace{-0.5cm}
\end{figure}

Fig.~\ref{fig:proposed-method} shows the elements of our approach.
We use the Gaussian splatting approach proposed in SplaTAM~\cite{keetha2024splatam} to generate maps.
We propose an efficient information gain metric and use a hierarchical 
planning framework to plan \emph{high-level} navigation targets that yield maximal information in the environment and \emph{low-level} paths that are \emph{dynamically feasible}, \emph{collision-free} and \emph{maximize the local information of the path}.
We utilize 3DGS as a unified representation for both mapping and planning. We avoid the cost of maintaining additional volumetric representations -- achieving the same granularity of a 3DGS map with a voxel map would require around 19.2 GB of RAM for a $40m \times 40 m \times 3 m$ space at 1 $cm$ resolution. With our GPU-based trajectory planner, we achieve an 18x speedup compared to a CPU-based implementation, as detailed in Sec.~\ref{subsec:planner_performance}. This allows us to generate collision-free trajectories on millions of Gaussians efficiently.
Our system runs real-time onboard a fully autonomous unmanned ground vehicle (UGV) to explore an unknown environment while generating a high-fidelity visual representation of the environment.

In summary, the contributions of this paper are:
\begin{enumerate}
    \item A \textbf{unified framework for online mapping and planning} built only on Gaussian splatting, eliminating the need for maintaining additional volumetric representations.
    \item An information gain heuristic that is easy to compute and capable of running in \textbf{real-time onboard a robot}.
    \item A real-time exploration system built on Gaussian splatting, \textbf{comprehensively validated in both simulation and real-world experiments across diverse indoor and outdoor environments.} We will release the full autonomy framework as open-source.
\end{enumerate}
    
\section{Related Work}
\label{sec:related work}
\textbf{Map Representation.} To effectively construct a map of the environment, numerous map representations have been proposed in the robotics community. The most intuitive but effective volumetric representation has been widely used. Voxel-based representation could maintain information such as occupancy~\cite{hornung2013octomap} or signed distance~\cite{Voxblox}. 
With the recent application of semantic segmentation, semantic maps that contain additional task-relevant information have been proposed~\cite{hughes2022hydra, 10057106, liu2024slideslamsparselightweightdecentralized}. With the recent advances in learned map representations in the computer vision community, Neural Radiance Fields (NeRF)~\cite{mildenhall2021nerf} and 3D Gaussian Splatting (3DGS)~\cite{kerbl20233dgs} have become popular representations for robotic motion planning~\cite{he2023active, chen2024splat, jin2024gsplanner}. 
In this work, we study the problem of active mapping with the 3DGS representation.

\textbf{Active Mapping.}
The problem of exploration and active \gls{slam} has been widely studied in the past decade. The classical exploration framework uses a model-based approach to actively navigate towards frontiers~\cite{yamauchi1997frontier} or waypoints that have the highest information gain~\cite{LukasIG, alexis2020MP,he2023active}. 
Some recent work combines the idea of frontier exploration and the information-driven approach to further improve efficiency~\cite{zhou2021fuel, yuezhantao2023seer, ral24_active_3d_slam}. However, most of the existing work developed their approaches based on classical map representations. 
In this work, we instead consider the active mapping problem with a 3DGS representation.
Bayesian neural networks~\cite{pan2022activenerf, lee2024bayesian} and deep ensembles~\cite{he2023active} are common approaches for estimating uncertainty in learned representations.
Radiance field representations provide additional possibilities for estimating uncertainty through the volumetric rendering process~\cite{Lee2022nerf3drecon}. NARUTO~\cite{feng2024naruto} uses an uncertainty-learning module to quantify uncertainty for active reconstruction.
CG-SLAM~\cite{hu2024cgslam} uses the difference between rendered depth and observations, and the alignment of $\alpha$-blended and median depth as measures of uncertainty.
FisherRF~\cite{jiang2023fisherrfactiveviewselection} leverages Fisher information to compute pixel-wise uncertainty on rendered images.
Where prior work in implicit and radiance field representations use indirect methods like ensembles and rendering for estimating uncertainty in the representation, a map represented by Gaussians encodes physical parameters of the scene, which motivates estimating information from the Gaussian parameters directly without rendering.

\textbf{Planning in Radiance Fields.}
Prior work has also considered planning directly in radiance fields.
GS-planner~\cite{jin2024gsplanner} plans trajectories in a Gaussian map and uses observability coverage and reconstruction loss stored in a voxel grid as an approximation of information gain for exploration.
Sim-to-real approaches leverage radiance fields as a simulator for imitation learning~\cite{murali2024av}.
Adamkiewicz et al.~\cite{adamkiewicz2022vision} utilize the geometric fidelity of the representation to first map the environment and then perform trajectory optimization to plan paths in these environments.
Splat-nav~\cite{chen2024splat} uses a pre-generated Gaussian map to compute safe polytopes to generate paths.
The probabilistic representations of free space has also been utilized for motion planning \cite{chen2024catnips} where authors use uncertainties in the learned representation to provide guarantees on collision-free paths.
While prior work considers planning in radiance fields, they typically maintain a separate volumetric representation for planning which requires additional memory usage onboard the robot.
On the other hand, maintaining a secondary representation may lead to inconsistencies in collision checks for planning. 
Existing work that performs collision checking with 3DGS requires subsampling to provide soft constraints. In this work, our proposed planning module performs dense checks directly against all Gaussians for collision avoidance.

\section{Problem Specification \& Preliminaries}
\label{sec:prelim}

\begin{figure*}[!t]
        \centering
        \vspace{0.1cm}
        \includegraphics[width=0.87\linewidth,clip]{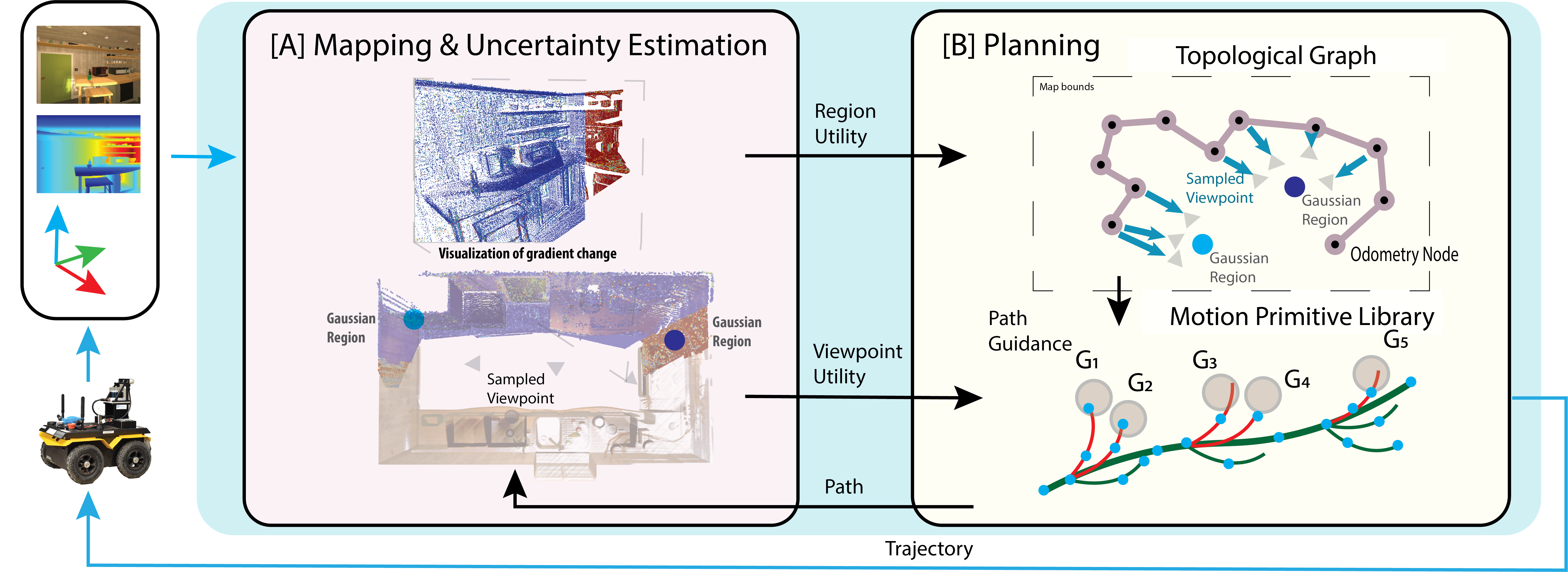}
        \caption{The proposed active mapping framework. The proposed framework contains two major components, the planning module and the mapping module. 
        As can be seen in the figure, the Mapping module ([A]) takes in RGB, depth and pose measurements, and updates the map representation at every step and computes the utility of cuboidal regions (Sec.~\ref{sec:high-level-guidance}).
        The utility of each region is then passed to the planning module which comprises the topological graph and motion primitive library ([B]). 
        \WIP{Viewpoints are sampled around regions in the map and added to the topological graph. A high-level guidance path is generated from the topological graph to viewpoints with highest utility and passed to the trajectory planner.}
        The trajectory planner in turn attempts to plan a path to goal that maximizes information gathering (queried from the mapper).
        The planned trajectory is executed by the robot to get a new set of observations.
       } 
\label{fig:system}
\vspace{-0.5cm}
\end{figure*}

Our goal is to construct an estimate $\hat{G}$ of the true map of the environment $G^{*}$, which is \textit{a priori} unknown. 
The quality of the constructed map $\hat{G}$ is evaluated using a hidden test set $\mathcal{T}$ of tuples of poses and corresponding noise-free measurements, $(x_{test}^{*}, y_{test}^{*})_{test \in \mathcal{T}}$. 
We aim to solve the following problem
\begin{equation}
\begin{aligned}
        \argmin_{\hat{G}, \ x_{1:T}} & \sum_{{test}\in \mathcal{T}} \mathcal{L}(y^*_{test} , h(x^*_{test}, \hat{G}))\\
        s.t & \  \hat{G} = \Phi(x_{1:T}, y_{1:T}), \\
\end{aligned}
\label{eq:true_problem}
\end{equation}
where $\mathcal{L}(\cdot)$ captures the difference between a synthesized and a true measurement, $\Phi(\cdot)$ is the function that constructs $\hat{G}$ from the set of measurements $y_{1:T}$ obtained at poses $x_{1:T}$, $h(\cdot)$ is the rendering function that synthesizes a measurement given a pose and a map, and $T$ is the time budget for exploration.

The problem above is ill-posed, as the set of test poses and measurements $\mathcal{T}$ is hidden. 
We approach this challenge by solving the following two problems in a receding horizon scheme. 
At time $k \leq T$, we first solve a mapping problem that involves minimizing the difference between the observed measurements $y_{1:k}$ and synthesized measurements $\hat{y}_{1:k}$: 
\begin{equation}
\Phi(x_{1:k}, y_{1:k}) :=  
\argmin_{\hat{G}}  \sum_{s = 1}^{k} \mathcal{L}(y_{s} , h(x_{s}, \hat{G})).
\label{eq:mapping_problem}
\end{equation}
Given the posterior estimate of the map $\hat{G}$, we find the subsequent viewpoint $x_{k+1}$ that maximizes the information gain $\mathcal{I}(\cdot)$ given by the mutual information between the map $\hat{G}$ and the measurement $y_{k+1}$ at the corresponding pose: 
\begin{equation}
    \max_{x_{k+1}} \ \mathcal{I} (\hat{G}; y_{k+1} \mid x_{k+1}).
\end{equation}

\noindent \textbf{Gaussian Splatting.} We use 3D Gaussian Splatting (3DGS)~\cite{kerbl20233dgs} to represent the environment as a collection of isotropic Gaussians $\hat{G}$. 
Each Gaussian is parameterized with 8 values representing RGB intensities ($c$), 3D position ($\mu$), radius ($r$), and opacity ($\alpha$). 
To generate the measurement at a given camera pose of the estimated map $\hat{G}$, 
3D Gaussians are sorted in increasing order of depth relative to the camera pose and are then splatted onto the image plane.
In particular, each splatted Gaussian will have a projected 2D mean ($\mu_{2D}$) and radius ($r_{2D}$) on the image plane,
\[
\scalemath{0.9}{
\mu_{2D} = \frac{K R_{k}^\intercal (\mu - T_{k})}{d}, \ r_{2D} = \frac{fr}{d}, \  d = e_3^\intercal R_{k}^\intercal  (\mu - T_{k})},
\]
where $K$ is the intrinsic matrix, $R_{k}$ and $T_{k}$ are the rotational and translational components of the camera pose in the world frame at time $k$, $f$ is the focal length, and $e_3 = [0,0,1]^\intercal$.

For a given pixel $p$ in the image, its color $\mathcal{C(\cdot)}$ can be obtained as
\begin{equation*}
\scalemath{0.85}{
\mathcal{C}(p) = \sum_{i=1}^{n} c_i f_i(p) \prod_{j=1}^{i-1} (1-f_j(p)), f_i(p) = \alpha_i \exp \left\{ -\frac{||p-\mu_{2D, i}||^2}{2r_{2D, i}^2}\right\}}.
\end{equation*}
Similarly, the depth $\mathcal{D(\cdot)}$ can be obtained by replacing $c_i$ with $d_i$.
At every iteration, we render color image $\hat{y}_{c}$ and depth image $\hat{y}_{d}$. We set the function $\mathcal{L}$ (from eq. \eqref{eq:mapping_problem}) to
\begin{equation}
\scalemath{0.850}{\mathcal{L} = \frac{1}{N_{p}}\sum_{p\in y} \left( \vert \hat{y}_d - y_{d}\vert + \lambda_1  \vert \hat{y}_c - y_{c} \vert \right) + \lambda_2(1-SSIM(\hat{y}_{c}, y_{c}))}
\end{equation}
which is a weighted combination of the L1 loss on depth and rendered pixel colors and the structured similarity index measure (SSIM) to update the parameters of the Gaussians, and $N_p$ is the number of pixels in the image.

\section{Method}

Our proposed framework comprises a mapping module and a planning module, as illustrated in Fig.~\ref{fig:system}. 
The mapping module (Sec.~\ref{sec:mapping}) accumulates measurements and poses to reconstruct the environment and computes the uncertainty of Gaussians in the map. 
This is then passed to the planning module for planning guidance paths (Sec.~\ref{sec:high-level-guidance}) and trajectories (Sec.~\ref{sec:mpl}).

\subsection{Mapping \& Uncertainty Estimation}
\label{sec:mapping}
We build the 3DGS mapping module upon SplaTAM~\cite{keetha2024splatam} with isotropic Gaussians. 
As in \cite{keetha2024splatam}, we set $\lambda_1 = 0.4, \lambda_2=0.1$.
At the beginning of each mapping iteration, we initialize new Gaussians at positions corresponding to the current color and depth measurements. 
The parameters of these Gaussians are refined through backpropagation. 
As in \cite{kerbl20233dgs}, we prune Gaussians with low opacity or large radius.

\noindent \textbf{Uncertainty Estimation.} 
The objective for finding the next-best-view can then be expressed as maximizing the mutual information
\begin{equation}
\begin{aligned}
\scalemath{0.9}{x^*_{k+1}} & \scalemath{0.9}{= \argmax_{x_{k+1}} \mathcal{I} (y_{k+1}; \hat{G} \mid x_{k+1})},  \\
     &\scalemath{0.9}{\doteq {H} (y_{k+1} \mid x_{k+1} ) - {H} (y_{k+1}  \mid \hat{G}, x_{k+1})}
\end{aligned}
\end{equation}
where ${H}(\cdot)$ is the Shannon entropy. We have
\[
\scalemath{0.9}{
H (y_{k+1} \mid x_{k+1} ) = - \int p(y_{k+1} \mid x_{k+1}) \log p(y_{k+1} \mid x_{k+1})) d y_{k+1}}
\]
with $\scalemath{0.9}{p(y_{k+1} \mid x_{k+1}) = \int p(y_{k+1} \mid \hat{G}, x_{k+1}) p(\hat{G}) d\hat{G}}$ and
\begin{align*}
\scalebox{0.9}{$H(y_{k+1} \mid$} & \scalebox{0.9}{$\hat{G}, x_{k+1}) =-\int p(\hat{G})$} \\
   &\scalebox{0.9}{$\biggl[ \int p(y_{k+1}  \mid \hat{G}, x_{k+1})
       \log p(y_{k+1} \mid \hat{G}, x_{k+1}) \, dy_{k+1} \biggr] d\hat{G}$}
\end{align*}
We make the following assumptions in this work to approximate the information gain efficiently: (a)
$\hat{G} \sim \mathcal{N} (\E [G], \Var [G])$, where $G$ is the random variable that denotes the map; (b) $p(y_{k+1} \mid \hat{G}) = \mathcal{N} (h(x_{k+1}, \hat{G}), \Sigma_y)$; with (c) constant isotropic variance $\Sigma_y$. 
For the marginal entropy ${H} (y_{k+1} \mid x_{k+1})$, given assumptions (a) and (b), and linearization of $h(\cdot)$ around $\E[G]$, we can approximate the distribution $p(y \mid x) = \mathcal{N}(h(x,\E [G]), J_{k+1}\Var [G] J_{k+1}^\intercal)$, where $J_{k+1} = \partial h(x_{k+1}, \hat{G})/\partial G$. Under assumptions (b) and (c), the conditional entropy does not depend on $G$.
Recall the entropy of a $d_Z$ dimensional Gaussian distribution $\mathcal{N} (\mu_Z, \Sigma_Z)$ is $\frac{d_Z}{2} \log (2\pi e) + \frac{1}{2} \log \det (\Sigma_Z)$. 
Dropping the constant terms, the information gain can be simplified to
\begin{equation}
\scalemath{0.9}{\mathcal{I}(y_{{k+1}} ; \hat{G} \mid x_{k+1} ) 
=\frac{1}{2} \log \det (I + \Sigma_y^{-1} J_{k+1} \Var [G] J_{k+1}^\intercal )}.
\end{equation}
Using the first order approximation of $\log \det$ about the identity, we can define a proxy metric for scoring next best view as
\begin{equation}
\scalemath{0.9}{
    x^*_{k+1} = \argmax_{x_{k+1}} \mathrm{Tr} (\Sigma_y^{-1} J_{k+1} \Var [G] J_{k+1}^\intercal)}.
\label{eq:nbv}
\end{equation}
As in~\cite{jiang2023fisherrfactiveviewselection}, assumption (c) allows us to simplify the optimization by ignoring the effect of $\Sigma^{-1}_y$.
Evaluating $J_{k+1}$ is computationally expensive because it typically requires rendering at new poses. 
To achieve real-time evaluation of multiple candidate viewpoints, 
we replace the Jacobian term $J_{k+1}$ with a binary matrix that corresponds to its sparsity pattern. Intuitively, this captures the set of Gaussians within the field of view at pose $x_{k+1}$. 
The key insight behind these approximations is that we estimate information gain using the current uncertainty, based on the assumption that local measurements lead to precise local map estimates.
We show empirically that the exploration performance based on our metric (in Sec.~\ref{subsubsection:ai2thor_benchmark}) is similar to the performance based on eq.~\eqref{eq:nbv}.
We propose a heuristic for estimating uncertainty $\Var [G]$ based on the magnitude of the change in means ($\mu$) of the Gaussians, associating larger displacements with higher uncertainty. 
The motivation behind this is that $\Var [G]$ can be approximated with the empirical Fisher Information matrix~\cite{martens2020new}, $\Var [G] \approx \nabla_{G} \mathcal{L}(x_k, y_k)\nabla_{G} \mathcal{L}(x_k, y_k)^\intercal$, which in turn can be directly linked to the square of the updates of Gaussian parameters tuned by the gradient descent algorithm. 
In practice, we observe that
the square of the magnitude of the updates outlined above 
is sensitive to measurement noise (in Sec.~\ref{subsubsection:ai2thor_ablation}), so we use the magnitude of the change instead. 

Empirically, we observe that the means of Gaussians located near boundaries between observed and unobserved space exhibit significant changes over successive updates, indicating high uncertainty, as visualized in Fig.~\ref{fig:system}. Consequently, directing exploration toward these high-uncertainty Gaussians is analogous to following frontiers in grid-based exploration, making this heuristic effective for autonomous exploration.
Through careful bookkeeping, Gaussians that remain unchanged in the most recent update retain their previously computed displacement values, ensuring that our heuristic provides a consistent measure of each Gaussian's movement over time. In addition to the boundaries between observed and unobserved space, areas within observed space that have insufficient measurements also exhibit relatively high uncertainty. This further directs the robot to explore these areas and gather more information, improving the overall map quality.

\subsection{Hierarchical Planning}
In spite of the approximations introduced in the previous section to compute information gain obtained by navigating to a given area in space in a computationally efficient way, the task of finding such a region is still a challenging optimization problem. For this reason, we address the latter task through a hierarchical planning framework. 
The high-level planner provides guidance to map particular regions of the environment and a path to that region from the known space in the environment.
The low-level (trajectory) planner finds a trajectory that is dynamically-feasible (\ie obeys the robot's physical constraints), collision-free and locally maximizes the information gain along the path.

\subsubsection{High-level planner} 
\label{sec:high-level-guidance}
In contrast to traditional mapping representations, the Gaussian map does not encode occluded and free space.
Instead of computing (geometric) frontiers, we use the Gaussian uncertainty estimates to identify regions of the map that should be visited next. 
We evenly partition the \textit{a priori} known enclosing space of the desired map into cuboidal regions.
At time ${k}$, for a region $o$, denote the cardinality of the Gaussians in the region by $M_{k}$.   
We compute the mean uncertainty of that region
$
\Omega_{k} = 1/{M_{k}} \left(\sum_{\mu \in o} \| \mu_{k} - \mu_{{k-1}} \|_2\right).
$

We formulate a high-level guidance path that allows us to (i) navigate to regions of high uncertainty; and (ii) utilize the known traversable space to plan long-range trajectories without computationally expensive collision checks.
At the high level of our planner, we construct a tree by incrementally adding nodes along the traveled path.
The tree consists of two types of nodes: \emph{odometry} nodes and \emph{viewpoint} nodes. As the robot moves, it verifies the presence of a nearby \emph{odometry} node; if none exist, a new \emph{odometry} node is added to the tree and connected to the nearest existing \emph{odometry} nodes.
At each planning iteration, we sample a fixed number of viewpoints around the identified regions.
We compute the shortest viewing distance from the Gaussian region centroid
to the optical center of the camera, given the camera intrinsics.
Each sampled viewpoint is then assigned the utility computed for that region and connected to the closest \emph{odometry} node in the tree as a \emph{viewpoint} node.
We then use Dijkstra's algorithm to find the shortest path from the current robot location in the tree to all the \emph{viewpoint} nodes in the tree.
Finally, we compute the cost-benefit of a path using $\Omega / e^{d}$ where $\Omega$ is the estimated information gain and $d$ is the distance to the node~\cite{topo2020indoor}.
The maximal cost-benefit path is sent to the trajectory planner.

\subsubsection{Trajectory planner}
\label{sec:mpl} 

We partition the current map $\hat{G}$ into three disjoint subsets: $\hat{G}_H$, which consists of Gaussians with high uncertainty, $\hat{G}_L$, which consists of Gaussians with low uncertainty, and $\hat{G}_O$ for the rest.
For a Gaussian $g \in \hat{G}$ and pose $x$, we define the binary visibility function $v(x, g)$ capturing whether $g$ is in the field of view of $x$.
To avoid rendering each view to check for occlusions, we cull the Gaussians that are beyond the perception range of the sensor.

To maximize information gain while exploring the environment, we aim to obtain viewpoints with high utility $\xi$ that maximize the number of high-uncertainty Gaussians and minimize the number of low-uncertainty Gaussians in the field of view.
Trajectories are evaluated based on the sum of the utilities of the viewpoints in each trajectory  given by
\begin{equation}
\scalemath{0.9}{
  \xi(x, \hat{G}) = \sum_{g \in \hat{G}_H} v(x, g) - \lambda_{\xi} \sum_{g \in \hat{G}_L} v(x, g)}.
  \label{equ:view_util}
\end{equation}
The first term in eq.~\eqref{equ:view_util} encourages additional observations of parts of the scene with high uncertainty. The second term penalizes observation of stable Gaussians in explored areas and encourages exploration of unseen parts of the environment.
The two terms are weighted by $\lambda_{\xi}$.

Given the path generated by the high-level planner, we select the furthest point $p_{goal}$ along the path that lies within a local planning range $R_H$, and set it as the center of the goal region $\mathcal{X}_{goal}$.
We use the unicycle model as the robot dynamics $f(\cdot)$. 
The robot state $x=[p, \theta] \in \mathbb{R}^{2} \times S^{1}$ consists of its position ($p$) and heading ($\theta$). The control inputs $u = [v, \omega] \in \mathbb{R}^2$ consist of linear velocity ($v$) and angular velocity $(\omega)$.
We solve the trajectory planning problem in 2 steps: (i) finding feasible, collision-free trajectory candidates, and (ii) selecting a trajectory that maximizes the information gain.
The first problem is defined as follows:
\begin{problem}
\textbf{Trajectory Candidate Generation.}
Given an initial robot state $x_{0} \in \mathcal{X}_{free}$, and goal region $\mathcal{X}_{goal}$, find the control inputs $u(\cdot)$ defined on $[0,\tau]$ that solve:
\begin{equation}
\begin{aligned}
    & \min_{u(\cdot), \tau}  \ \lambda_{t} \tau + \int_{0}^\tau u(t)^Tu(t)  dt \\
    & s.t. \ \forall  t  \in [0,\tau], \ x(0) = x_0,  \  x(\tau) \in \mathcal{X}_{goal},\\ 
        &\dot{x}(t) = f(x(t), u(t)) , \  x(t) \in \mathcal{X}_{free},\\ 
        &|| v(t) ||_2 \leq v_{max},  \  | \omega(t) | \leq \omega_{max}\\ %
\end{aligned}
\label{equ: low_level_planner}
\end{equation}
where $\lambda_t$ weights the time cost with the control effort, $v_{max}$ and $\omega_{max}$ are actuation constraints, and $x_{0}$ is the initial state.  
\label{prob:low-level-planning}
\end{problem}

\noindent \textbf{Motion Primitive Tree Generation.} Inspired by~\cite{liu2017mpl}, we solve problem~\ref{prob:low-level-planning} by performing a tree search on the motion primitives tree. 
Motion primitives are generated with fixed control inputs over a time interval with known initial states. 
Given the actuation constraints $v_{max}$ and $\omega_{max}$ of the robot, we uniformly generate $N_{v} \times N_{\omega}$ samples from $[0, v_{max}] \times [-\omega_{max}, \omega_{max}]$ as the finite set of control inputs. 
Subsequently, motion primitives are constructed given the dynamics model, the controls $u$, and time discretization. 
\noindent \textbf{Collision Checking.}
We sample a fixed number of points on each motion primitive to conduct collision checks. 
Since we have an uncertain map, we relax $x \in X_{free}$ to a chance constraint.
Let $d_\kappa$ be the minimum distance between a test point and the set of Gaussians.
Our constraint then amounts to the probability of the distance between the test point and the set of Gaussians in the scene being less than some allowed tolerance $\gamma$ with probability $\eta$ \ie $P(d_\kappa < \gamma) \leq \eta$.
When checking for collisions, each sampled point is bounded with a sphere of radius $r_{robot}$ and the radius $r$ of each Gaussian is scaled by a factor of $\lambda_{g}$.
The truth value of the test is determined by comparing the distance to all Gaussians with $\gamma := r_{robot} + \lambda_{g} r$, as illustrated in Fig.~\ref{fig:system}. 
Note that setting $\lambda_g=3$ is equivalent to the test proposed by~\cite{jin2024gsplanner}. However, unlike the continuous trajectory optimization approach in \cite{jin2024gsplanner}, which requires subsampling to provide soft constraints, our search-based method allows dense checks.
We are implicitly assuming that 3DGS accurately captures geometry of the scene and we notice this holds empirically since we initialize Gaussians in the map using the depth measurements from the RGB-D sensor. 
In the case of collision checking against non-isotropic Gaussians, the principal axis of the Gaussian can be taken as the radius.
We develop a GPU-accelerated approach for testing collisions of all sampled points with all Gaussians from the map at once while growing the search tree. 
This allows the real-time expansion of the search tree.

\noindent \textbf{Tree Search.}
The cost of each valid motion primitive is defined by Prob.~\ref{prob:low-level-planning}.
Since the maximum velocity of the robot is bounded by $v_{max}$, we consider the minimum time heuristic as $h(p) := ||p_{goal} - p||_{2}/v_{max}$.
We use A* to search through the motion primitives tree and keep top $N_{traj}$ candidate trajectories for the information gain maximization. 

\noindent \textbf{Information Maximization.}
A state sequence $\{x_{i}({l \tau /L}) \ \vert \ l \in \{0, \ldots,L\}\}$ containing the end state of each of $L$ segments in the $i$-th trajectory is used for evaluating the information along the trajectory.
The information of each state is evaluated according to eq.~\eqref{equ:view_util}.
Finally, the trajectory with the highest information 
\begin{equation}
\argmax_{i \in \{1, \ldots, N_{traj}\}} \sum_{l=0}^{L} \xi \left(x_{i}\left(\frac{l\tau}{L} \right) \right), \ l\in \{0, \ldots, L\}
\label{equ:traj_utility}
\end{equation}
is selected and executed by the robot. 

\subsection{Implementation Details}
We implemented the proposed method using PyTorch2.4 and the mapping module following~\cite{keetha2024splatam}.
We note that our framework is compatible with other Gaussian splatting approaches if they can meet the compute and latency requirements. 
The parameters of the mapper were set to the default configuration of~\cite{keetha2024splatam} for the TUM dataset.
For viewpoint selection experiments, we used 50 mapping iterations.
For real-world experiments, to enable real-time mapping and collision avoidance, we reduced the number of mapping iterations to 10 and pruned Gaussians once per mapping sequence. The mapping and planning modules are both set to 1Hz onboard the robot.
For trajectory viewpoint evaluation, we set the threshold for high and low uncertainty Gaussians at half a standard deviation above and below the average magnitude of parameter updates, respectively. We found Gaussians on the ground plane to be particularly noisy in indoor environments due to reflections off the floor.
This necessitated the removal of Gaussians on the ground plane for planning.
We set the Gaussian region size to 2.5m, $\lambda_{\xi} = 1$ and $\lambda_{g} = 3$. For planning, we set $N_v = 3$, $v_{max}=0.6$, $N_\omega = 5$, $\omega_{max}= 0.9$, and time discretization 1s.

\section{Results}
\label{sec:results}

\subsection{Information Metric Evaluation \& Ablation}
\label{subsec:ai2thor_sim}

\subsubsection{Experiment setup \& Baselines}
\label{subsubsec:sim_setup_baseline_metrics}

To evaluate our proposed heuristic on viewpoint selection, we conducted simulation experiments with the AI2-THOR simulator~\cite{ai2thor} on the iTHOR dataset. In these experiments, time budget is not imposed, and we run experiments on all test scenes with 100 steps each. The iTHOR dataset consists of indoor scenes including kitchens, bedrooms, bathrooms and living rooms. 
At each step, the agent is allowed to translate 0.5m in both $x$ and $y$ positions and a discrete set of possible orientations.
These candidate positions, along with their corresponding yaw angles, define the full set of potential viewpoints.
Each viewpoint is evaluated using an uncertainty metric, and the one with the highest uncertainty is selected as the next action.
The simulations were run on a desktop computer with an AMD Ryzen Threadripper PRO 5975WX and  NVIDIA RTX A4000 (16GB).

\begin{table}[!t]
\centering
\vspace{0.17cm}
\caption{\textbf{Information metric benchmark experiments on iTHOR}}
\begin{tabular}{c|c|c|c|c|c}
\hline
\multirow{2}{*} {\textbf{Methods}}& \textbf{PSNR}  & \textbf{SSIM}  & \textbf{LPIPS}  & \textbf{RMSE}  & \textbf{t/step}  \\
& [dB] $\uparrow$ & $\uparrow$ & $\downarrow$ & [m] $\downarrow$ & [s] $\downarrow$ \\
\hline
Ensemble & 21.489 & 0.762 & 0.360 & 0.350 & 2.667 \\
FisherRF\textsuperscript{\textdagger} & \cellcolor{red!40}\textbf{24.171} & \cellcolor{red!40}\textbf{0.849} & \cellcolor{red!40}\textbf{0.270} & \cellcolor{red!40}\textbf{0.285} & \cellcolor{orange!40}1.299 \\
\textit{RTGuIDE} & \cellcolor{orange!40}22.946 & \cellcolor{orange!40}0.820 & \cellcolor{orange!40}0.305 & \cellcolor{orange!40}{0.343} & \cellcolor{red!40}\textbf{0.013} \\ 
\hline
\multicolumn{6}{p{0.41\textwidth}}{\textsuperscript{\textdagger} Viewpoint selected using information metric from FisherRF.}
\end{tabular}
\label{table:benchmark_ithor_results}
\end{table}

\begin{table}[!t]
\centering
\caption{\textbf{Information metric ablation experiments on iTHOR}}
\begin{tabular}{c|c|c|c|c|c}
\hline
\multirow{2}{*} {\textbf{Methods}}& \textbf{PSNR}  & \textbf{SSIM}  & \textbf{LPIPS}  & \textbf{RMSE}  & \textbf{t/step}  \\
& [dB] $\uparrow$ & $\uparrow$ & $\downarrow$ & [m] $\downarrow$ & [s] $\downarrow$ \\
\hline
\textit{RTGuIDE} & \textbf{22.946} & \textbf{0.820} & \textbf{0.305} & \textbf{0.343} & {0.013} \\ 
{RTGuIDE-sum} & 20.279 & 0.740 & 0.371 & 0.512 & \textbf{0.007}\\ 
{RTGuIDE-sq} & 22.489 & 0.802 & 0.308 & {0.343} & 0.014\\ 
{RTGuIDE w. noise} & 22.541 & 0.801 & 0.329 & 0.344 & 0.013\\ 
{RTGuIDE-sq w. noise} & 21.445 & 0.779 & 0.351 & 0.386 & 0.013\\ 
\hline
\end{tabular}
\label{table:ablation}
\vspace{-0.4cm}
\end{table}

For each scene, we synthesized novel views from the generated maps from a set of uniformly sampled test poses and computed the image metrics against the ground truth images. 
We evaluate the Peak Signal-to-Noise Ratio (PSNR), Structural Similarity Index (SSIM)~\cite{1292216}, Learned Perceptual Image Patch Similarity (LPIPS)~\cite{zhang2018perceptual} on the RGB images and Root Mean-Square-Error (RMSE) on the depth images.
We evaluate the proposed method (denoted as \textit{Ours}) on viewpoints following eq.~\eqref{equ:view_util} against two other view selection methods.
For \textit{FisherRF}, we follow the original implementation to evaluate viewpoints based on the Fisher information.
For the \textit{Ensemble} baseline, we train an ensemble of $5$ models with the leave-one-out procedure at each step. We compute the patch-wise variance across the rendered images for each sampled viewpoint and ensemble model, and select the viewpoint with the highest variance as the next action. 

\subsubsection{Benchmark experiments}
\label{subsubsection:ai2thor_benchmark}
The averaged results over all test scenes are presented in Tab.~\ref{table:benchmark_ithor_results}.
Our heuristic performs comparably with \textit{FisherRF} and 
outperforms \textit{Ensemble} across all metrics, while being more than an order of magnitude faster in computation time.
\textit{FisherRF} renders images at every sampled viewpoint to approximate the information gain with the full map parameters, while we avoid such computational cost by computing the uncertainty directly on the parameters of the Gaussians. This trades off the approximation accuracy with efficiency.
This efficiency in computation is crucial for real-time operation where we evaluate potentially hundreds of viewpoints in each planning iteration.

\subsubsection{Ablation experiments}
\label{subsubsection:ai2thor_ablation}
For ablation experiments, we implemented \texttt{RTGuIDE-sum}, which solely considers the sum of the uncertainty of Gaussians in the camera view. \texttt{RTGuIDE-sq} evaluates the uncertainty of Gaussians as the squared L2 norm following the original derivation in eq.~\eqref{eq:nbv} and using eq.~\eqref{equ:view_util} to evaluate the uncertainty of viewpoints. 
To simulate real-world noisy measurements, we added Gaussian noise $(\mu=0, \sigma^2=0.1)$ to depth measurements in the simulation to evaluate \texttt{RTGuIDE w. noise} and \texttt{RTGuIDE-sq w. noise}.

As shown in Tab.~\ref{table:ablation}, the ablation experiments on different metrics demonstrate that our proposed approach outperforms the simple uncertainty summation ({RTGuIDE-sum}), which focuses purely on exploitation. Furthermore, the results show that both the squared ({RTGuIDE-sq}) and standard L2 norms ({RTGuIDE}) achieve comparable performance when perfect depth measurements are available. However, when noise is introduced, the experiments confirm that our chosen metric is more robust and better suited for real-world scenarios where measurement noise is inevitable.

\subsection{Planner Performance}
\label{subsec:planner_performance}
\begin{figure}
\centering
\vspace{0.05cm}
\begin{tikzpicture}
  \pgfmathsetlengthmacro\MajorTickLength{
    \pgfkeysvalueof{/pgfplots/major tick length} * 0.05
  }
  \pgfplotsset{every tick label/.append style={font=\footnotesize}}
  \begin{axis}[
      width=0.48\textwidth,
      height=4.5cm,
      ylabel={Time (ms)},
      xlabel={Number of Gaussians},
      xmode=log,
      major tick length=\MajorTickLength,
      grid=both,
      xlabel style={font=\footnotesize, yshift=3pt}, 
      ylabel style={font=\footnotesize, yshift=-9pt}, 
      tick label style={font=\tiny},
      legend style={font=\scriptsize},
      grid style={line width=.1pt, draw=gray!10},
      major grid style={line width=.2pt,draw=gray!50},
      every tick/.style={black, semithick},
      legend pos=north west,
  ]



\addplot+[blue, thick, dashed, mark=o, mark size=1.5pt, mark options={fill=blue, draw=blue, solid}]
    table[x index=0, y index=1, col sep=comma] {collision-test-time.dat};
\addlegendentry{MPL-CPU}
\addplot+[blue, only marks, mark=o, mark size=1.5pt, mark options={fill=blue, draw=blue, solid}, forget plot]
    plot[error bars/.cd, y dir=both, y explicit]
    table[x index=0, y index=1, y error index=2, col sep=comma] {collision-test-time.dat};

\pgfplotsset{cycle list shift=0}

  \addplot+[red, thick, solid, mark=square*, mark size=1.5pt, mark options={fill=red}]
      table[x index=0, y index=1, col sep=comma] {collision-test-time-2.dat};
  \addlegendentry{MPL-GPU}
  \addplot+[red, only marks, mark=square*, mark size=1.5pt, mark options={fill=red}, forget plot]
      plot[error bars/.cd, y dir=both, y explicit]
      table[x index=0, y index=1, y error index=2, col sep=comma] {collision-test-time-2.dat};

  \addlegendentry{Desired latency}
  \addplot[mark=none, black, thick, dashed] coordinates {(100000,1000) (5000000,1000)};
  \end{axis}
\end{tikzpicture}
\vspace{-1.0cm}
\caption{Time spent to plan a trajectory with a 5m horizon (including 129 collision checking points) versus the number of Gaussians in the map.}
\label{fig:planning_time}
\end{figure}
We evaluate the necessity of our proposed GPU-based planning approach in enabling real-time planning. 
In particular, we conducted experiments on trajectory planning with a 5-meter horizon, evaluating both GPU-accelerated collision checking and a fully CPU-based implementation. 
As illustrated in Fig.~\ref{fig:planning_time}, the planning time of CPU-based planner increases significantly with the number of Gaussians. For example, a 40m $\times$ 40m outdoor parking lot can contain around $3\times10^6$ Gaussians. In this case, the CPU-based planner requires 11.65 seconds, whereas the GPU-based planner takes 0.62 seconds, achieving 18$\times$ speedup. The GPU-based collision testing enables parallel growth of the search tree at each layer, along with simultaneous and efficient collision checks between all test points and Gaussians directly on the GPU.

\subsection{Closed-loop Simulation Experiments}
\label{subsec:sim_experiment}

\begin{table}[!t]
\centering
\vspace{-0.2cm}
\caption{\textbf{Closed-loop Experiments on MP3D in Unity Simulator}}
\begin{tabular}{c|c|c|c|c|c}
\hline
\multirow{2}{*} {\textbf{Methods}}& \textbf{PSNR}  & \textbf{SSIM}  & \textbf{LPIPS}  & \textbf{RMSE}  & \textbf{Coverage}  \\
& [dB] $\uparrow$ & $\uparrow$ & $\downarrow$ & [m] $\downarrow$ & [\%] $\uparrow$ \\
\hline
Ensemble & 10.71 & 0.24 & 0.76 & 2.14 & 38.93 \\
FisherRF & \cellcolor{orange!40}15.37 & \cellcolor{orange!40}0.49 & \cellcolor{orange!40}0.59 & \cellcolor{orange!40}1.06 & \cellcolor{orange!40}66.40 \\
\textit{RTGuIDE} & \cellcolor{red!40}\textbf{16.32} & \cellcolor{red!40}\textbf{0.56} & \cellcolor{red!40}\textbf{0.54} & \cellcolor{red!40}\textbf{0.71} & \cellcolor{red!40}\textbf{92.36} \\
\cdashline{1-6}
{GT} & 18.13 & 0.65 & 0.47 & 0.50 & -- \\ 
\hline
\end{tabular}
\label{table:simulation_unity}
\vspace{-0.5cm}
\end{table}

\begin{table*}[!h]
\centering
\vspace{0.17cm}
\caption{\textbf{Quantitative Results of Real-world Experiments.}
}
\begin{tabular}{c|c|c|c|c|c|c|c|c}
\hline
 {\textbf{Methods}}& {\textbf{Env.}}  & \textbf{Budget (Time)} [s] & \textbf{PSNR} [dB] $\uparrow$ & \textbf{SSIM} $\uparrow$ & \textbf{LPIPS} $\downarrow$ & \textbf{RMSE} [m] $\downarrow$ & \textbf{Coverage} [\%] $\uparrow$ & \textbf{mIoU}$^{\mathrm{a}}$  $\uparrow$  \\
\hline
Ensemble    & \multirow{4}{*}{\makecell{Indoor 1 \\ 72m$^2$}} & \multirow{4}{*}{{300}} & 6.27 & 0.204 & 0.839  & 2.051 & 98.48 & 0.071 \\
FisherRF  & & & 16.80 & 0.665 & 0.393 & 0.242 & 99.86 & 0.314 \\
\textit{RTGuIDE} &  &  & \textbf{17.83} & \textbf{0.737} & \textbf{0.334} & \textbf{0.202}  & 99.95 & \textbf{0.338} \\ \cdashline{1-1} \cdashline{4-9}
GT & &  & 20.56 & 0.805 & 0.237 & 0.158 & -- & 0.420 \\
\hline
Ensemble    & \multirow{4}{*}{\makecell{Indoor 2 \\ 189m$^2$}} & \multirow{4}{*}{{600}} & 10.41 & 0.269 & 0.787 & 2.410  & 84.10 & 0.156  \\
FisherRF  & & & 14.88 & 0.588 & 0.451 & 1.170  & 72.80 & 0.248 \\
\textit{RTGuIDE} &  &  & \textbf{16.40} & \textbf{0.711} & \textbf{0.299} & \textbf{0.689}  & 100.00 & \textbf{0.391} \\ \cdashline{1-1} \cdashline{4-9}
GT & &  & 18.39 & 0.803 & 0.243 & 0.454 & -- & 0.422 \\ 
\hline
Ensemble    & \multirow{4}{*}{\makecell{Outdoor 1 \\ 252m$^2$}} & \multirow{4}{*}{{600}} & 10.59 & 0.392 & 0.637 & 1.705  & 48.30 & 0.549 \\
FisherRF  & & & 17.58 & 0.687 & 0.349 & 0.519  & 85.81 & 0.600 \\
\textit{RTGuIDE} &  &  & \textbf{20.00} & \textbf{0.744} & \textbf{0.296} & \textbf{0.369}  & 100.00 & \textbf{0.622} \\ \cdashline{1-1} \cdashline{4-9}
GT & &  & 22.95 & 0.828 & 0.241 & 0.195 & -- & 0.631 \\ 
\hline
Ensemble    & \multirow{4}{*}{\makecell{Outdoor 2 \\ 480m$^2$}} & \multirow{4}{*}{{900}} & 16.79 & 0.538 & 0.576 & 1.422  & 49.59 & 0.159 \\
FisherRF  & & & 19.44 & 0.680 & 0.439 & 0.773  & 60.95 & \textbf{0.177} \\
\textit{RTGuIDE} &  &  & \textbf{20.27} & \textbf{0.750} & \textbf{0.345} & \textbf{0.399}  & 98.35 & 0.174 \\ \cdashline{1-1} \cdashline{4-9}
GT & &  & 24.19 & 0.835  & 0.285 & 0.227 & -- & 0.246 \\
\hline
\multicolumn{9}{p{0.9\textwidth}}{$^{\mathrm{a}}$ Indoor 1 classes: [floor, chair, table, refrigerator, cabinet, backpack, plants]. Indoor 2 classes: [floor, chair, table, couch, cushion, trashcan, television, plants]. Outdoor 1 classes: [pavement, plants, barrel, traffic cone]. Outdoor 2 classes: [pavement, grass, tree, bench, lamppost].} \\
\end{tabular}
\label{tab:realworld_results}
\vspace{-0.65cm}
\end{table*}

\subsubsection{Experiment setup \& Baselines}
We conducted experiments in a Unity-based simulator with ROS and a simulated Clearpath Jackal robot with an RGBD sensor and ground truth odometry. This set of experiments was designed to evaluate all methods with real-time and closed-loop operation, considering robot dynamics and collision avoidance.

The original implementation of \textit{FisherRF}~\cite{jiang2023fisherrfactiveviewselection} does not model dynamics or include real-world experimental results, so we implement a closed-loop version of it.
We first construct both a voxel map and a Gaussian splatting map online.
We detect and cluster frontiers from the voxel map and use A* to generate reference paths to frontier clusters.
Camera views along each path are evaluated to compute information gain. 
To generate collision-free and dynamically-feasible trajectories, we generate trajectories along waypoints in the path through MPL~\cite{liu2017mpl} in the voxel map.
We similarly implement an \textit{Ensemble} method as described in Sec.~\ref{subsubsec:sim_setup_baseline_metrics}.

We obtained the ground truth (GT) by tele-operating the robot to uniformly survey the environment.
The collected data was used to build a 3DGS map for GT and also served as test poses for the other methods.
This represents the upper bound of the evaluation metrics with the Gaussian splatting method.
The exploration budget for all methods is set to approximately 5 times the duration needed to fully survey the environment with tele-operation. 
For the baseline methods, we set the voxel resolution to 5cm.
We rendered 500 novel views at the ground truth test poses for evaluation. 
The same image metrics on the rendered RGB and depth images are used as in Sec.~\ref{subsec:ai2thor_sim}.
We evaluated all methods on 5 MP3D~\cite{Matterport3D} scenes \texttt{[17DRP5sb8fy, HxpKQynjfin, 2t7WUuJeko7, 8194nk5LbLH, YVUC4YcDtcY]} with a time budget of 1500s, 480s, 1500s, 1200s and 1800s respectively. These scenes consist of multi-room indoor environments with varying size and clutter.

\subsubsection{Results}
The averaged statistics of the experiments are presented in Table~\ref{table:simulation_unity}. We note that there are signifcant errors in the meshes in MP3D which affect the reconstruction results for all methods including the ground truth. Our proposed method achieves 0.95dB and 5.61dB higher PSNR than FisherRF and Ensemble. The significantly higher coverage demonstrates the importance of efficient viewpoint selection in real-time missions with limited operational budget.

\subsection{Real-world Experiments \& Benchmarks}
\label{subsec:real_world_experiment}

\subsubsection{Experiment setup \& Baselines}
To evaluate the effectiveness of our proposed framework in real-world settings, we compared it with two baselines onboard the robot.
We followed the same setup as in the closed-loop simulation experiments for the baselines and the ground truth GT.

We deployed our method and the baselines in four different real-world environments on a Clearpath Jackal robot outfitted with an AMD Ryzen 5 3600 and RTX 4000 Ada SFF.
In addition, the platform is equipped with an Ouster OS1 LiDAR for state estimation and a ZED 2i stereo camera for mapping. 
We used \cite{9718203} to provide odometry and truncated the depth measurements at 5 meters for all methods in all of our experiments.
The same evaluation setup from Sec.~\ref{subsec:sim_experiment} is employed.
To verify the usefulness of the generated Gaussian map representation for downstream tasks in robotics, we also performed an evaluation on the task of semantic segmentation. We used Grounded SAM 2~\cite{ren2024grounded} to obtain segmentation masks of the rendered and groundtruth images and computed the mean Intersection over Union (mIoU).

\subsubsection{Qualitative results}

\begin{figure} [!t]
\centering
\includegraphics[width=0.98\linewidth, trim={0.0cm 0.0cm 0cm 0cm},clip]{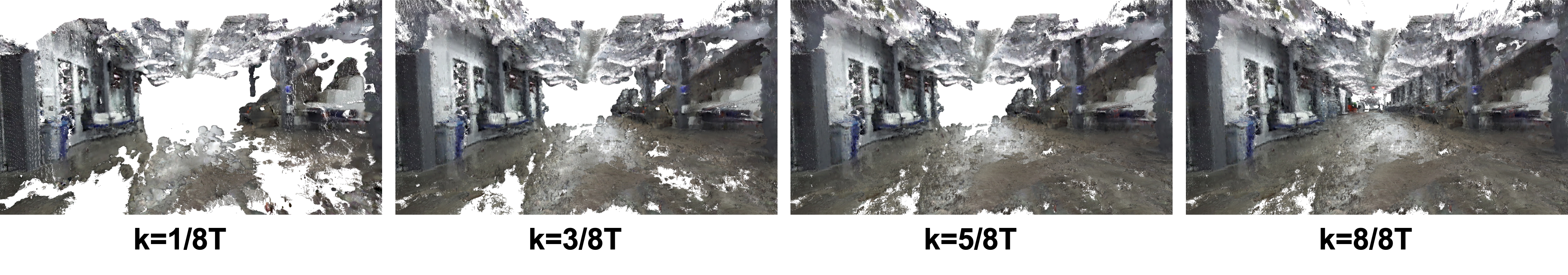}
\vspace{-0.2cm}
\caption{Visualization of onboard constructed map}
\label{fig:indoor_maps}
\vspace{-0.4cm}
\end{figure}

\begin{figure} [!t]
\centering
\includegraphics[width=0.98\linewidth, trim={0.0cm 0cm 0cm 0cm},clip]{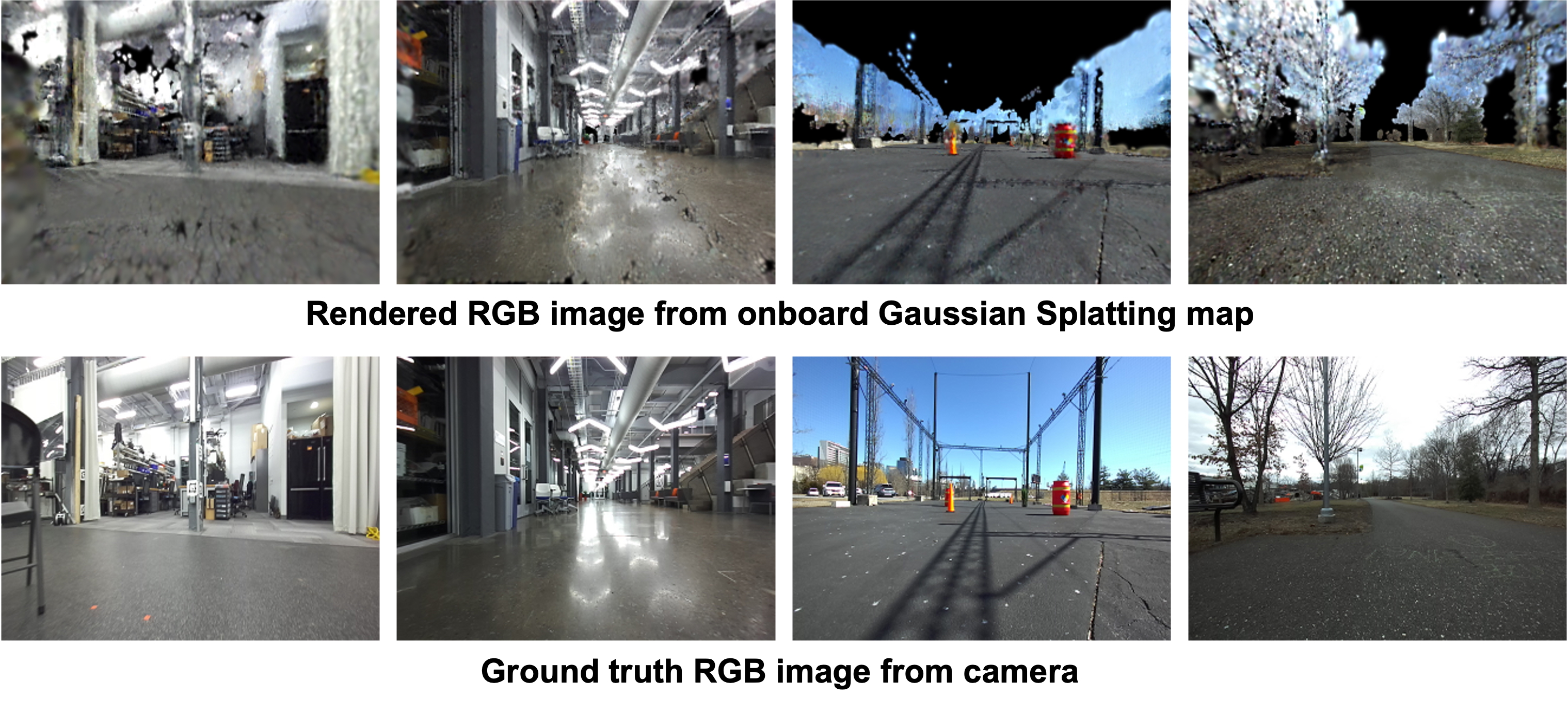}
\vspace{-0.2cm}
\caption{Qualitative results}
\label{fig:qualitative}
\vspace{-0.7cm}
\end{figure}

We evaluated the quality of the Gaussian splatting maps that were constructed in real-time onboard the robot by rendering novel views with a set of test poses.
A visualization of the onboard constructed map is shown in Fig.~\ref{fig:indoor_maps}, and examples of rendered color images are presented in Fig.~\ref{fig:qualitative}.
Constructed in real time onboard the robot, our Gaussian splatting map provides a detailed representation of the environment.
Being able to render photorealistic images from the map benefits downstream tasks such as semantic segmentation, which are analyzed in the next section.

\subsubsection{Quantitative results}
As shown in Tab.~\ref{tab:realworld_results}, in real-world scenarios when the operational budget is limited, our proposed framework constructs maps of higher quality than the baselines onboard the robot.
To validate that our approach generates maps of reasonable quality, we also present maps generated with the ground truth samples.
Results on novel view image rendering show that the maps constructed with our approach achieve 0.8-2.4 higher PSNR than FisherRF and 3.5-11.6 higher PSNR than Ensemble.
Furthermore, when evaluated on depth reconstruction, our method reduces the RMSE of the rendered depth by 16.5\%-48\% compared to FisherRF, and by 71.4\%-90.2\% compared to Ensemble.
These results reflect that our method outperforms the baselines in achieving good coverage of the environments while also optimizing for map quality during exploration.

For semantic segmentation, our approach achieves better mIoU scores for most experiments, indicating higher fidelity of the rendered images compared to the baselines.

\subsubsection{Discussion}
We attribute the performance of our approach to two key factors. First, the use of an information metric that is efficient to compute onboard the robot enables smooth and continuous operation. Second, since our method does not rely on frontiers extracted from the voxel map for geometric coverage, it allows for revisiting of areas in the environment to further improve map quality.

\section{Limitations and Future work}
\label{sec:conclusion}
In this work, we present a framework for real-time active exploration and mapping with Gaussian splatting.
A future direction is to consider semantic features together with the Gaussian splatting maps to perform complex tasks in the environment like object search and represent dynamic scenes in our framework.  
In this work, we study the active mapping problem which assumes perfect state estimation. 
In future work, we aim to incorporate the state estimation uncertainty and formulate this entire framework using a single sensor.  

\bibliography{abbrev_literature}

\begin{thebibliography}{10}
\providecommand{\url}[1]{#1}
\csname url@rmstyle\endcsname
\providecommand{\newblock}{\relax}
\providecommand{\bibinfo}[2]{#2}
\providecommand\BIBentrySTDinterwordspacing{\spaceskip=0pt\relax}
\providecommand\BIBentryALTinterwordstretchfactor{4}
\providecommand\BIBentryALTinterwordspacing{\spaceskip=\fontdimen2\font plus
\BIBentryALTinterwordstretchfactor\fontdimen3\font minus \fontdimen4\font\relax}
\providecommand\BIBforeignlanguage[2]{{%
\expandafter\ifx\csname l@#1\endcsname\relax
\typeout{** WARNING: IEEEtran.bst: No hyphenation pattern has been}%
\typeout{** loaded for the language `#1'. Using the pattern for}%
\typeout{** the default language instead.}%
\else
\language=\csname l@#1\endcsname
\fi
#2}}

\bibitem{mildenhall2021nerf}
B.~Mildenhall, P.~P. Srinivasan, M.~Tancik, J.~T. Barron, R.~Ramamoorthi, and R.~Ng, ``Nerf: Representing scenes as neural radiance fields for view synthesis,'' \emph{Commun. ACM}, vol.~65, no.~1, pp. 99--106, 2021.

\bibitem{kerbl20233dgs}
B.~Kerbl, G.~Kopanas, T.~Leimk{\"u}hler, and G.~Drettakis, ``3d gaussian splatting for real-time radiance field rendering,'' \emph{ACM Trans. Graph.}, vol.~42, no.~4, pp. 1--14, 2023.

\bibitem{LukasIG}
L.~{Schmid}, M.~{Pantic}, R.~{Khanna}, L.~{Ott}, R.~{Siegwart}, and J.~{Nieto}, ``An efficient sampling-based method for online informative path planning in unknown environments,'' \emph{IEEE Robot. Automat. Lett.}, vol.~5, no.~2, pp. 1500--1507, 2020.

\bibitem{alexis2020MP}
M.~Dharmadhikari, T.~Dang, L.~Solanka, J.~Loje, H.~Nguyen, N.~Khedekar, and K.~Alexis, ``Motion primitives-based path planning for fast and agile exploration using aerial robots,'' in \emph{Proc. IEEE Int. Conf. Robot. Automat.}, 2020, pp. 179--185.

\bibitem{yamauchi1997frontier}
B.~Yamauchi, ``A frontier-based approach for autonomous exploration,'' in \emph{Proc. IEEE Int. Symp. Comput. Intell. Robot. Automat.}\hskip 1em plus 0.5em minus 0.4em\relax IEEE, 1997, pp. 146--151.

\bibitem{10143748}
J.~Yu, H.~Shen, J.~Xu, and T.~Zhang, ``Echo: An efficient heuristic viewpoint determination method on frontier-based autonomous exploration for quadrotors,'' \emph{IEEE Robot. Automat. Lett.}, vol.~8, no.~8, pp. 5047--5054, 2023.

\bibitem{he2023active}
H.~Siming, C.~D. Hsu, D.~Ong, Y.~S. Shao, and P.~Chaudhari, ``Active perception using neural radiance fields,'' in \emph{2024 American Control Conference (ACC)}.\hskip 1em plus 0.5em minus 0.4em\relax IEEE, 2024, pp. 4353--4358.

\bibitem{keetha2024splatam}
N.~Keetha, J.~Karhade, K.~M. Jatavallabhula, G.~Yang, S.~Scherer, D.~Ramanan, and J.~Luiten, ``Splatam: Splat, track \& map 3d gaussians for dense rgb-d slam,'' in \emph{Proceedings of the IEEE/CVF Conference on Computer Vision and Pattern Recognition}, 2024.

\bibitem{hornung2013octomap}
A.~Hornung, K.~M. Wurm, M.~Bennewitz, C.~Stachniss, and W.~Burgard, ``Octomap: An efficient probabilistic 3d mapping framework based on octrees,'' \emph{Auton. Robots}, vol.~34, pp. 189--206, 2013.

\bibitem{Voxblox}
H.~{Oleynikova}, Z.~{Taylor}, M.~{Fehr}, R.~{Siegwart}, and J.~{Nieto}, ``Voxblox: Incremental 3d euclidean signed distance fields for on-board mav planning,'' in \emph{Proc. IEEE/RSJ Int. Conf. Intell. Robots Syst.}, 2017, pp. 1366--1373.

\bibitem{hughes2022hydra}
N.~Hughes, Y.~Chang, and L.~Carlone, ``Hydra: A real-time spatial perception system for {3D} scene graph construction and optimization,'' \emph{Robot.: Sci. Syst.}, 2022.

\bibitem{10057106}
A.~Asgharivaskasi and N.~Atanasov, ``Semantic octree mapping and shannon mutual information computation for robot exploration,'' \emph{IEEE Trans. Robot.}, vol.~39, no.~3, pp. 1910--1928, 2023.

\bibitem{liu2024slideslamsparselightweightdecentralized}
X.~Liu, J.~Lei, A.~Prabhu, Y.~Tao, I.~Spasojevic, P.~Chaudhari, N.~Atanasov, and V.~Kumar, ``Slideslam: Sparse, lightweight, decentralized metric-semantic slam for multi-robot navigation,'' \emph{arXiv preprint arXiv:2406.17249}, 2024.

\bibitem{chen2024splat}
T.~Chen, O.~Shorinwa, J.~Bruno, A.~Swann, J.~Yu, W.~Zeng, K.~Nagami, P.~Dames, and M.~Schwager, ``Splat-nav: Safe real-time robot navigation in gaussian splatting maps,'' \emph{IEEE Trans. Robot.}, 2025.

\bibitem{jin2024gsplanner}
R.~Jin, Y.~Gao, Y.~Wang, Y.~Wu, H.~Lu, C.~Xu, and F.~Gao, ``Gs-planner: A gaussian-splatting-based planning framework for active high-fidelity reconstruction,'' in \emph{Proc. IEEE/RSJ Int. Conf. Intell. Robots Syst.}\hskip 1em plus 0.5em minus 0.4em\relax IEEE, 2024, pp. 11\,202--11\,209.

\bibitem{zhou2021fuel}
B.~Zhou, Y.~Zhang, X.~Chen, and S.~Shen, ``Fuel: Fast uav exploration using incremental frontier structure and hierarchical planning,'' \emph{IEEE Robot. Automat. Lett.}, vol.~6, no.~2, pp. 779--786, 2021.

\bibitem{yuezhantao2023seer}
Y.~Tao, Y.~Wu, B.~Li, F.~Cladera, A.~Zhou, D.~Thakur, and V.~Kumar, ``{SEER}: Safe efficient exploration for aerial robots using learning to predict information gain,'' in \emph{Proc. IEEE Int. Conf. Robot. Automat.}\hskip 1em plus 0.5em minus 0.4em\relax IEEE, 2023, pp. 1235--1241.

\bibitem{ral24_active_3d_slam}
Y.~Tao, X.~Liu, I.~Spasojevic, S.~Agarwal, and V.~Kumar, ``3d active metric-semantic slam,'' \emph{IEEE Robot. Automat. Lett.}, vol.~9, no.~3, pp. 2989--2996, 2024.

\bibitem{pan2022activenerf}
X.~Pan, Z.~Lai, S.~Song, and G.~Huang, ``Activenerf: Learning where to see with uncertainty estimation,'' in \emph{Proc. Eur. Conf. Comput. Vision}.\hskip 1em plus 0.5em minus 0.4em\relax Springer, 2022, pp. 230--246.

\bibitem{lee2024bayesian}
S.~Lee, K.~Kang, and H.~Yu, ``Bayesian nerf: Quantifying uncertainty with volume density in neural radiance fields,'' \emph{arXiv preprint arXiv:2404.06727}, 2024.

\bibitem{Lee2022nerf3drecon}
S.~Lee, L.~Chen, J.~Wang, A.~Liniger, S.~Kumar, and F.~Yu, ``Uncertainty guided policy for active robotic 3d reconstruction using neural radiance fields,'' \emph{IEEE Robot. Automat. Lett.}, vol.~7, no.~4, pp. 12\,070--12\,077, 2022.

\bibitem{feng2024naruto}
Z.~Feng, H.~Zhan, Z.~Chen, Q.~Yan, X.~Xu, C.~Cai, B.~Li, Q.~Zhu, and Y.~Xu, ``Naruto: Neural active reconstruction from uncertain target observations,'' in \emph{Proc. IEEE Conf. Comput. Vis. Pattern Recognit.}, 2024, pp. 21\,572--21\,583.

\bibitem{hu2024cgslam}
J.~Hu, X.~Chen, B.~Feng, G.~Li, L.~Yang, H.~Bao, G.~Zhang, and Z.~Cui, ``Cg-slam: Efficient dense rgb-d slam in a consistent uncertainty-aware 3d gaussian field,'' in \emph{Proc. Eur. Conf. Comput. Vision}.\hskip 1em plus 0.5em minus 0.4em\relax Springer, 2024, pp. 93--112.

\bibitem{jiang2023fisherrfactiveviewselection}
W.~Jiang, B.~Lei, and K.~Daniilidis, ``Fisherrf: Active view selection and mapping with radiance fields using fisher information,'' in \emph{Proc. Eur. Conf. Comput. Vision}.\hskip 1em plus 0.5em minus 0.4em\relax Springer, 2024, p. 422–440.

\bibitem{murali2024av}
V.~Murali, G.~Rosman, S.~Karamn, and D.~Rus, ``Learning autonomous driving from aerial views,'' in \emph{Proc. IEEE/RSJ Int. Conf. Intell. Robots Syst.}\hskip 1em plus 0.5em minus 0.4em\relax IEEE, 2024.

\bibitem{adamkiewicz2022vision}
M.~Adamkiewicz, T.~Chen, A.~Caccavale, R.~Gardner, P.~Culbertson, J.~Bohg, and M.~Schwager, ``Vision-only robot navigation in a neural radiance world,'' \emph{IEEE Robot. Automat. Lett.}, vol.~7, no.~2, pp. 4606--4613, 2022.

\bibitem{chen2024catnips}
T.~Chen, P.~Culbertson, and M.~Schwager, ``Catnips: Collision avoidance through neural implicit probabilistic scenes,'' \emph{IEEE Trans. Robot.}, vol.~40, pp. 2712--2728, 2024.

\bibitem{martens2020new}
J.~Martens, ``New insights and perspectives on the natural gradient method,'' \emph{J. Mach. Learn. Res.}, vol.~21, no. 146, pp. 1--76, 2020.

\bibitem{topo2020indoor}
C.~Gomez, M.~Fehr, A.~Millane, A.~C. Hernandez, J.~Nieto, R.~Barber, and R.~Siegwart, ``Hybrid topological and 3d dense mapping through autonomous exploration for large indoor environments,'' in \emph{Proc. IEEE Int. Conf. Robot. Automat.}, 2020, pp. 9673--9679.

\bibitem{liu2017mpl}
S.~Liu, N.~Atanasov, K.~Mohta, and V.~Kumar, ``Search-based motion planning for quadrotors using linear quadratic minimum time control,'' in \emph{Proc. IEEE/RSJ Int. Conf. Intell. Robots Syst.}\hskip 1em plus 0.5em minus 0.4em\relax IEEE, 2017, pp. 2872--2879.

\bibitem{ai2thor}
E.~Kolve, R.~Mottaghi, W.~Han, E.~VanderBilt, L.~Weihs, A.~Herrasti, M.~Deitke, K.~Ehsani, D.~Gordon, Y.~Zhu, \emph{et~al.}, ``Ai2-thor: An interactive 3d environment for visual ai,'' \emph{arXiv preprint arXiv:1712.05474}, 2017.

\bibitem{1292216}
Z.~Wang, E.~Simoncelli, and A.~Bovik, ``Multiscale structural similarity for image quality assessment,'' in \emph{The Thrity-Seventh Asilomar Conference on Signals, Systems \& Computers, 2003}, vol.~2, 2003, pp. 1398--1402 Vol.2.

\bibitem{zhang2018perceptual}
R.~Zhang, P.~Isola, A.~A. Efros, E.~Shechtman, and O.~Wang, ``The unreasonable effectiveness of deep features as a perceptual metric,'' in \emph{Proc. IEEE Conf. Comput. Vis. Pattern Recognit.}, 2018.

\bibitem{Matterport3D}
A.~Chang, A.~Dai, T.~Funkhouser, M.~Halber, M.~Niessner, M.~Savva, S.~Song, A.~Zeng, and Y.~Zhang, ``Matterport3d: Learning from rgb-d data in indoor environments,'' \emph{Int. Conf. 3D Vision}, 2017.

\bibitem{9718203}
C.~Bai, T.~Xiao, Y.~Chen, H.~Wang, F.~Zhang, and X.~Gao, ``Faster-lio: Lightweight tightly coupled lidar-inertial odometry using parallel sparse incremental voxels,'' \emph{IEEE Robot. Automat. Lett.}, vol.~7, no.~2, pp. 4861--4868, 2022.

\bibitem{ren2024grounded}
T.~Ren, S.~Liu, A.~Zeng, J.~Lin, K.~Li, H.~Cao, J.~Chen, X.~Huang, Y.~Chen, F.~Yan, Z.~Zeng, H.~Zhang, F.~Li, J.~Yang, H.~Li, Q.~Jiang, and L.~Zhang, ``Grounded sam: Assembling open-world models for diverse visual tasks,'' 2024.

\end{thebibliography}

\end{document}